# Monotonicity and Persistence in Preferential Logics

**Joeri Engelfriet**  JOERI@CS.VU.NL
*Vrije Universiteit, Faculty of Mathematics and Computer Science*
*De Boelelaan 1081a, 1081 HV Amsterdam, The Netherlands*

## Abstract

An important characteristic of many logics for Artificial Intelligence is their nonmonotonicity. This means that adding a formula to the premises can invalidate some of the consequences. There may, however, exist formulae that can always be safely added to the premises without destroying any of the consequences: we say they respect monotonicity. Also, there may be formulae that, when they are a consequence, can not be invalidated when adding any formula to the premises: we call them conservative. We study these two classes of formulae for preferential logics, and show that they are closely linked to the formulae whose truth-value is preserved along the (preferential) ordering. We will consider some preferential logics for illustration, and prove syntactic characterization results for them. The results in this paper may improve the efficiency of theorem provers for preferential logics.

## 1. Introduction

Over the past decades, many non-classical logics for Artificial Intelligence have been defined and investigated. The need for such logics arose from the unsuitability of classical logics to describe defeasible reasoning. These classical logics are monotonic, which means that their consequence relation ($\mid\sim$) satisfies:

$$\forall \alpha, \beta, \varphi : (\alpha \mid\sim \beta \Rightarrow \alpha \wedge \varphi \mid\sim \beta) \qquad (Monotonicity)$$

This means that whenever we learn new information ($\varphi$) and add this to what we already know ($\alpha$), all the old theorems ($\beta$) are still derivable. This is clearly undesirable when describing defeasible reasoning. Therefore, monotonicity is not satisfied by many logics for Artificial Intelligence.

On the other hand, monotonicity is a very attractive feature from a practical point of view. When learning new information, we do not have to start all over again, but we can retain our old conclusions, and focus on deriving possible new ones. Furthermore, when we have a lot of information, we are allowed to focus on only part of it. Conclusions derived from this part are then automatically also valid when considering all the information we have (this is sometimes called *local reasoning*).

Even though it is clear that we do not want monotonicity to hold in general, it might be worthwhile to investigate restricted variants of monotonicity. In the past, such variants have been defined which allow us to keep the old theorems, when either the new information follows from the old premise (this variant is called Cautious Monotonicity by Kraus, Lehmann, & Magidor, 1990) or its negation can not be derived from the old premise (this is called Rational Monotonicity by Kraus et al., 1990).

We will take a somewhat different perspective, and consider two classes of formulae: the class of formulae that can always be added to a premise without invalidating old conclusions (we say these formulae *respect monotonicity*), and the class of formulae which can always be retained as conclusions, no matter which new information is added to the premise (we say these formulae are *conservative*). The advantages of monotonicity sketched above would still hold when we restrict $\varphi$ to the class of formulae that respect monotonicity, or when we restrict $\beta$ to be conservative. Whether such classes exist, and what these classes are, depends of course on the particular nonmonotonic logic considered. We will focus here on an important class of nonmonotonic logics: the class of *preferential logics* (Shoham, 1987, 1988). These logics are based on a monotonic logic (such as propositional logic, predicate logic or modal logic) augmented with a preference order on its models. The nonmonotonic consequences of a formula $\alpha$ are those formulae which are true in all models of $\alpha$ which are minimal in the preference order among all models of $\alpha$ (an extensive discussion of preferential logics is provided by van Benthem, 1989). We will give a formal definition.

**Definition 1 (Preferential logic)** *A preferential logic consists of a language $\mathcal{L}$, a class of models Mod together with a satisfaction relation $\models$ between models and formulae, and a partial order $\preceq$ on Mod. A model $m \in Mod$ is called a* minimal model *of a formula $\alpha$ (denoted $m \models_{\preceq} \alpha$) if $m \models \alpha$ and for all models $n$, if $n \preceq m$ and $n \models \alpha$ then $n = m$.* Preferential entailment *($\models_{\preceq}$) between formulae is defined as follows: for $\alpha, \beta \in \mathcal{L} : \alpha \models_{\preceq} \beta$ if $\beta$ is true in all minimal models of $\alpha$.*

Our presentation uses a partial order, i.e., a reflexive, antisymmetric and transitive relation. Shoham (1987) uses a strict partial order, i.e., an irreflexive transitive relation, with a slightly different notion of minimal model. The presentations can be translated into each other.

It will turn out that formulae whose truth is preserved when going to more preferred or less preferred models, play an important role with respect to the two classes of formulae defined above (the class of formulae that respect monotonicity, and the class of conservative formulae). We will first give a definition.

**Definition 2 (Persistence)** *Given a preferential logic $(\mathcal{L}, Mod, \models, \preceq)$, a formula $\alpha \in \mathcal{L}$ is called* downward persistent *in this logic, if*

$$\forall m, n \in Mod : (m \models \alpha \text{ and } n \preceq m) \Rightarrow n \models \alpha,$$

*and it is called* upward persistent *if*

$$\forall m, n \in Mod : (n \models \alpha \text{ and } n \preceq m) \Rightarrow m \models \alpha.$$

In the next section, we will introduce some preferential logics to illustrate the material in the rest of the paper. In Section 3 we will consider formulae that respect monotonicity, and in Section 4 conservative formulae will be treated. The practical implications of the results of this paper are discussed in Section 5. Section 6 gives conclusions and suggestions for further research. Part of the material in this paper appeared in (Engelfriet, 1996b).

## 2. Some Preferential Logics

In this section we will describe the following preferential logics: Ground S5, Minimal Temporal Epistemic Logic and Circumscription. Since we have already defined preferential entailment in general, for each logic we only have to give its ingredients, i.e., $\mathcal{L}, Mod, \models$, and $\preceq$. The preferential entailment relation is then fixed by Definition 1. The first logic we will consider is Ground S5.

### 2.1 Ground S5

Ground S5 is a nonmonotonic modal logic for auto-epistemic reasoning, originally proposed by Halpern and Moses (1985). Their aim was to formalize statements of the form "I *only* know $\varphi$". It allows, for example, to derive that an agent which only knows $p$, does not know $q$. Ground S5 falls into the general scheme of ground nonmonotonic modal logics (Donini, Nardi, & Rosati, 1997). A lot of interest is devoted to logics of minimal knowledge (Levesque, 1990; Schwarz & Truszczyński, 1994; Chen, 1997; Halpern, 1997).

Semantically, states in which an agent only knows $\varphi$, are states in which $\varphi$ is known, but otherwise the amount of knowledge is minimal. We will use a modal propositional language to express the knowledge of the agent, and S5 will be the monotonic logic. We will give a treatment of Ground S5 slightly different, but equivalent to the one given by Halpern and Moses (1985).

**Definition 3 (Epistemic language)** *Let $P$ be a (finite or countably infinite) set of propositional atoms. The language $\mathcal{L}_{S5}$ is the smallest set closed under:*

- *if $p \in P$ then $p \in \mathcal{L}_{S5}$;*
- *if $\varphi, \psi \in \mathcal{L}_{S5}$ then $K\varphi, \varphi \wedge \psi, \neg \varphi \in \mathcal{L}_{S5}$.*

*Furthermore, we introduce the following abbreviations:*

$$\varphi \vee \psi \equiv \neg(\neg \varphi \wedge \neg \psi), \varphi \rightarrow \psi \equiv \neg \varphi \vee \psi, M\varphi \equiv \neg K \neg \varphi, \top \equiv p \vee \neg p, \bot \equiv \neg \top.$$

*If every atom occurring in a formula $\varphi$ is in the scope of a $K$ operator, we call $\varphi$ subjective.*

An example of a subjective formula is $\neg Kp \wedge K(q \rightarrow p)$, whereas $K(p \wedge q) \vee s$ is not subjective. In the rest of this paper we will only be interested in subjective formulae: they describe (just) the knowledge and ignorance of the agent.

In the usual S5 semantics, a model is a triple $(W, R, \pi)$, where $W$ is a set of worlds, $R$ is an equivalence relation on $W$ and $\pi$ is a function that assigns a propositional valuation to each world in $W$. We may however, in the case of one agent, restrict ourselves to *normal* S5 models, in which the relation is universal (each world is accessible from every world), and worlds are identified with propositional valuations (a proof of soundness and completeness of S5 with respect to these semantics is given by Meyer & van der Hoek, 1995).

**Definition 4 (S5 semantics)** *Let $P$ be a (finite or countably infinite) set of propositional atoms. A propositional valuation is a function from $P$ into $\{0, 1\}$ where $0$ stands for false and $1$ for true. The set of all such valuations will be denoted by $Mod(P)$. A normal S5* model $M$ is a non-empty subset of $Mod(P)$. The truth of an S5 formula $\varphi$ in such a model, evaluated in a world $m \in M$, denoted $(M, m) \models_{S5} \varphi$, is defined inductively:

1. $(M, m) \models_{S5} p \quad \Leftrightarrow \quad m(p) = 1$, for $p \in P$
2. $(M, m) \models_{S5} \varphi \wedge \psi \quad \Leftrightarrow \quad (M, m) \models_{S5} \varphi$ and $(M, m) \models_{S5} \psi$
3. $(M, m) \models_{S5} \neg\varphi \quad \Leftrightarrow \quad$ it is not the case that $(M, m) \models_{S5} \varphi$
4. $(M, m) \models_{S5} K\varphi \quad \Leftrightarrow \quad (M, m') \models_{S5} \varphi$ for every $m' \in M$

We have the following elementary results on subjective formulae. The proofs are straightforward.

**Proposition 5 (Subjective formulae)**

1. Let $\varphi$ be a subjective formula. For a normal S5 model $M$ and $m_1, m_2 \in M$ it holds:
   $$(M, m_1) \models_{S5} \varphi \Leftrightarrow (M, m_2) \models_{S5} \varphi.$$
   We define $M \models_{S5} \varphi$ if $(M, m) \models_{S5} \varphi$ for some, or, equivalently, all $m \in M$. The set of all normal S5 models, sometimes called *information states*, is denoted by IS.

2. An S5 formula $\varphi$ is subjective if and only if it is equivalent to a formula of the form $K\varphi$ with $\varphi \in \mathcal{L}_{S5}$.

A subjective formula describes the knowledge of an agent, but we want to formalize that this is *all* the agent knows. Therefore we are looking for models in which the knowledge of the agent is minimal, or in other words, in which the ignorance of the agent is maximal. We introduce a preference order over information states which favors models with less knowledge. The definition of this ordering is based on the observation that the more valuations the agent considers possible, the less knowledge the agent has. Indeed, for any propositional formula $\varphi$ we have: if $M_1 \models K\varphi$ and $M_1 \supseteq M_2$ then $M_2 \models K\varphi$.

**Definition 6 (Degree of knowledge)** *We define the* degree-of-knowledge *ordering $\preceq$ on normal S5 models as follows: for $M_1, M_2 \in IS : M_1 \preceq M_2 \Leftrightarrow M_1 \supseteq M_2$.*

Ground S5 is the preferential logic based on this ordering.

**Definition 7 (Ground S5)** *Ground S5 is the preferential logic with the subjective formulae of $\mathcal{L}_{S5}$ as its language, IS as its class of models, the satisfaction relation of Proposition 5 and the ordering of Definition 6. We will denote preferential entailment (as defined in Definition 1) of Ground S5 by $\models^{GS5}$.*

The reader can now check that, for instance, $Kp \models^{GS5} \neg Kq$. The (unique) minimal S5 model of $Kp$ consists of all propositional valuations in which $p$ is true, and this indeed contains a model in which $q$ is false. The entailment relation is nonmonotonic since $Kp \wedge Kq \not\models^{GS5} \neg Kq$. Another example illustrates the minimality of the agent's knowledge: $Kp \vee Kq \models^{GS5} \neg(Kp \wedge Kq)$.

Let us define a consequence relation $\hspace{0.5mm}\vert\hspace{-1mm}\sim\hspace{0.5mm}$ by $\varphi \hspace{0.5mm}\vert\hspace{-1mm}\sim\hspace{0.5mm} \psi$ if $K\varphi \models^{GS5} K\psi$. Then it turns out that this is the consequence relation of Halpern and Moses (1985), apart from the fact that they only defined it for premises which have a unique minimal model. Premises with a unique minimal model are called *honest*. To give an example, the formula $Kp$ is honest, but $Kp \vee Kq$ is not: both the S5 model consisting of all valuations in which $p$ is true, and the model with all valuations in which $q$ is true, are minimal models.

## 2.2 Minimal Temporal Epistemic Logic

In Ground S5, we are only able to express something about 'all the agent knows' at a particular instant. We will extend this logic in such a way that we are also able to say something about the changing knowledge of the agent over time. Originally, this extended logic was intended as a means of specifying nonmonotonic reasoning processes and of reasoning about their properties (Engelfriet & Treur, 1994, 1996; Engelfriet, 1996a). The idea is that a temporal formula describes the nonmonotonic inferences the agent has to perform during the course of a reasoning process (analogously to the use of temporal logic for specifying computer processes). But these inferences should also be the *only* cause of increases in the agent's knowledge. Therefore, we again have to minimize the agent's knowledge, but now over time (analogously to the use of minimization for dealing with the frame problem in temporal logics for describing action and change, Shoham, 1988). To this end, we will temporalize the epistemic language, epistemic models and the degree-of-knowledge ordering.

In order to describe past and future we introduce temporal operators $P, H, F, G$ and $\Box$, denoting respectively "sometimes in the past", "always in the past", "sometimes in the future", "always in the future" and "always". We do not want to describe the agent's knowledge of the future and past, but the future and past of the agent's knowledge. Therefore, temporal operators are not allowed to occur within the scope of the epistemic $K$ operator.

**Definition 8 (Temporal epistemic language)** *The language $\mathcal{L}_{\text{TEL}}$ is the smallest set closed under:*

- *if $\varphi \in \mathcal{L}_{\text{S5}}$ then $\varphi \in \mathcal{L}_{\text{TEL}}$;*
- *if $\alpha, \beta \in \mathcal{L}_{\text{TEL}}$ then $\alpha \wedge \beta, \neg\alpha, P\alpha, F\alpha \in \mathcal{L}_{\text{TEL}}$.*

*Again the abbreviations for $\vee, \rightarrow, \top$ and $\bot$ are introduced, as well as:*

$$G\alpha \equiv \neg F(\neg\alpha), H\alpha \equiv \neg P(\neg\alpha) \text{ and } \Box\alpha \equiv H\alpha \wedge \alpha \wedge G\alpha.$$

*If in the first clause we restrict ourselves to subjective formulae, we get the set of* subjective TEL formulae.

In the rest of this paper we will be interested in subjective TEL formulae since they describe how the knowledge of the agent is changing over time. Based on the set of natural numbers ($\mathbf{N}$) as flow of time, and normal S5 models as formalization of states in a temporal model, the following semantics is introduced for temporal epistemic logic (TEL):

**Definition 9 (Semantics of TEL)** *A TEL model is a function $\mathcal{M} : \mathbf{N} \to IS$. The truth of a formula $\varphi \in \mathcal{L}_{\text{TEL}}$ in $\mathcal{M}$ at time point $t \in \mathbf{N}$, denoted $(\mathcal{M}, t) \models \varphi$, is defined inductively as follows:*

1. $(\mathcal{M}, t) \models \varphi$   $\Leftrightarrow$   $\mathcal{M}(t) \models_{\text{S5}} \varphi$, if $\varphi \in \mathcal{L}_{\text{S5}}$
2. $(\mathcal{M}, t) \models \varphi \wedge \psi$   $\Leftrightarrow$   $(\mathcal{M}, t) \models \varphi$ and $(\mathcal{M}, t) \models \psi$
3. $(\mathcal{M}, t) \models \neg\varphi$   $\Leftrightarrow$   *it is not the case that* $(\mathcal{M}, t) \models \varphi$
4. $(\mathcal{M}, t) \models P\varphi$   $\Leftrightarrow$   $\exists s \in \mathbf{N}$ *such that* $s < t$ *and* $(\mathcal{M}, s) \models \varphi$
5. $(\mathcal{M}, t) \models F\varphi$   $\Leftrightarrow$   $\exists s \in \mathbf{N}$ *such that* $t < s$ *and* $(\mathcal{M}, s) \models \varphi$

A formula $\varphi$ is true in a model $\mathcal{M}$, denoted $\mathcal{M} \models \varphi$, if $(\mathcal{M}, 0) \models \varphi$. A TEL model $\mathcal{M}$ is called conservative (or a TELC model) if for all $s < t \in \mathbf{N} : \mathcal{M}(s) \preceq \mathcal{M}(t)$, with the ordering $\preceq$ of Definition 6. The set of TELC models is denoted by TCIS.

Note that the above definition is in principle ambiguous: a formula like $Kp \wedge Kq$ is an S5 formula which can be interpreted according to the first semantic clause, but it can also be seen as a conjunction, to be interpreted according to the second clause. As the interpretation of the conjunction (and the same holds for negation) is the same in S5 as in TEL, this ambiguity is harmless.

We will briefly explain the reason we have defined $\mathcal{M} \models \varphi$ if $(\mathcal{M}, 0) \models \varphi$. Later on, we will make the general assumption on preferential logics that the language contains negation, and that $m \models \neg \varphi$ if and only if $m \not\models \varphi$ (Assumption 1). If we define $\mathcal{M} \models \varphi$ if $(\mathcal{M}, t) \models \varphi$ for all $t \in \mathbf{N}$, then this assumption would not hold for TEL. The two definitions can be translated into each other, since $(\mathcal{M}, 0) \models \varphi$ if and only if $(\mathcal{M}, t) \models \neg P\top \rightarrow \varphi$ for all $t \in \mathbf{N}$ and $(\mathcal{M}, t) \models \varphi$ for all $t \in \mathbf{N}$ if and only if $(\mathcal{M}, 0) \models \varphi \wedge G\varphi$. This is also one of the reasons we gave a slightly different presentation of Ground S5 (using subjective formulae).

In conservative models, the propositional knowledge of an agent can only increase in time. We will restrict ourselves to these models, i.e., an agent can not forget or revise its (propositional) knowledge. Indeed, for a propositional formula $\varphi$ and a TELC model $\mathcal{M}$, if $(\mathcal{M}, t) \models K\varphi$ then $(\mathcal{M}, s) \models K\varphi$ for all $s > t$. This restriction can be made when the agent is reasoning about a fixed (non-changing) situation, and we are abstracting from particular implementation details (such as the use of backtracking implementations for nonmonotonic logics). We now extend the ordering and minimal consequence relation to TELC models. The ordering is extended in a pointwise fashion.

**Definition 10 (Minimal temporal epistemic logic)**

1. We extend the degree-of-knowledge ordering to TELC models by defining

$$\mathcal{M} \preceq \mathcal{N} \Leftrightarrow \text{ for all } s \in \mathbf{N} : \mathcal{M}(s) \preceq \mathcal{N}(s).$$

2. Minimal temporal epistemic logic (MTEL) is the preferential logic with the subjective TEL formulae as its language, TCIS as its class of models, the satisfaction relation ($\mathcal{M} \models \varphi$) of Definition 9 and the ordering of item 1. We will denote preferential entailment (as defined in Definition 1) of MTEL by $\models^{\text{MTEL}}$.

The idea behind using MTEL for specifying reasoning processes is that a subjective TEL formula $\varphi$ describes the reasoning of an agent over time (it can, for example, describe the use of nonmonotonic inference rules, see Proposition 11). The minimal models of $\varphi$ represent the process of the agent reasoning in time. We can then use minimal consequence to infer properties of this reasoning process.

It is easy to see that MTEL is a generalization of Ground S5: for subjective S5 formulae $\varphi, \psi$ we have that $\varphi \models^{\text{MTEL}} \psi$ if and only if $\varphi \models^{\text{GS5}} \psi$. As an example of the use of the notion of minimal temporal epistemic consequence, it has been shown by Engelfriet and Treur (1993) that it can capture default logic (Reiter, 1980b).

**Proposition 11 (Default logic in MTEL)** *Let a finite, propositional default theory $\Delta = \langle W, D \rangle$ be given and let*

$$\psi = \bigwedge \{\Box(K\alpha \wedge G(\neg K \neg \beta) \rightarrow G(K\gamma)) \mid (\alpha, \beta)/\gamma \in D\} \wedge \bigwedge \{K\alpha \mid \alpha \in W\}.$$

*Then $\varphi$ is a sceptical consequence of $\Delta$ in default logic if and only if $\psi \models^{\text{MTEL}} F(K\varphi)$.*

The conjuncts of the form $\Box(K\alpha \wedge G(\neg K \neg \beta) \rightarrow G(K\gamma))$ ensure the application of the default rules. In words: if the agent knows the prerequisite ($\alpha$), and the justification ($\beta$) remains consistent with what the agent knows throughout the future, then the agent must conclude the consequent ($\gamma$) in the next moment in time (and it will know $\gamma$ henceforth).

### 2.3 Circumscription

One of the earliest approaches to nonmonotonic reasoning is circumscription (McCarthy, 1977, 1980; Davis, 1980; Lifschitz, 1994; Etherington, 1988), a preferential logic based on first-order predicate logic. The main idea behind circumscription is a kind of completeness of information given to us: "the premises as stated give us 'the whole truth' about the matter" (van Benthem, 1989). This leads to at least two kinds of minimality: predicate-minimality and domain-minimality. The intuition behind predicate-minimality is that for some relevant property (predicate), all objects that have this property, are explicitly said to have this property in the premise. This allows us to formulate defaults stating that all normal objects have some property. Minimizing abnormality will allow us to conclude an object has this property, unless we can deduce from the premise that this object is abnormal. The intuition behind domain-minimality, is that the domain (of discourse) contains no other objects than those that can be deduced to exist from the premise. (This intuition is strongly tied to the domain-closure assumption of Reiter, 1980a). These two kinds of minimality are formalized by two variants of Circumscription. Both of them will be treated below.

The classical logic underlying circumscription is first-order predicate logic. We assume a standard first-order language $\mathcal{L}$ with a finite number of predicate symbols, including equality. We will also assume that the language contains no function or constant symbols. This is not a severe limitation, since we can eliminate function and constant symbols by introducing new predicate symbols (Davis, 1980). We will first give the definition of the orderings and then define predicate and domain circumscriptive consequence.

**Definition 12**

1. Let $P$ be a predicate symbol in the language $\mathcal{L}$. For a structure $M$ for the language, $P^M$ denotes the interpretation of $P$ in $M$ (so $P^M$ is a subset of $dom(M)^n$, where $dom(M)$ is the domain of $M$, and $n$ is the arity of $P$). For two structures $M, N$, we say $M$ is $P$-preferred to $N$, denoted $M \leq_P N$, if they have the same domain, the same interpretation of predicate symbols other than $P$, and $P^M \subseteq P^N$. Predicate circumscription of $P$ is the preferential logic which uses first-order predicate logic for the language, models and satisfaction relation, augmented with the ordering $\leq_P$. We will denote preferential entailment (as defined in Definition 1) in this logic by $\models_P^{\text{PC}}$.

2. For two structures $M, N$ for the language $\mathcal{L}$, we say $N$ is a substructure of $M$, denoted $N \leq_d M$, if the domain of $N$ is a subset of the domain of $M$, and the interpretation *of each predicate symbol in $N$ is the restriction of the corresponding interpretation in $M$ to $dom(N)$. Domain circumscription is the preferential logic which uses first-order predicate logic for the language, models and satisfaction relation, augmented with the ordering $\leq_d$. We will denote preferential entailment in this logic by $\models^{\text{DC}}$.*

3. *If we restrict the model class to finite structures, the resulting preferential logics are called* finite predicate circumscription *and* finite domain circumscription.

We refer the reader to the references given above for standard results and motivation of circumscription.

## 3. Respecting Monotonicity

In this section we will study formulae which respect monotonicity. We will first give a formal definition.

**Definition 13 (Respecting monotonicity)** *Given a preferential logic, we say a formula $\varphi$* respects monotonicity, *if*

$$\forall \alpha, \beta : \alpha \models_{\preceq} \beta \Rightarrow \alpha \wedge \varphi \models_{\preceq} \beta.$$

Next, we will make some basic assumptions about the (underlying logic of the) preferential logic.

**Assumption 1** *From now on we will assume that any preferential logic satisfies the following:*

- *the language has conjunction ($\wedge$) and $m \models \varphi \wedge \psi \Leftrightarrow m \models \varphi$ and $m \models \psi$.*

- *the language has implication ($\rightarrow$) and $m \models \varphi \rightarrow \psi \Leftrightarrow m \not\models \varphi$ or $m \models \psi$.*

- *the language has negation ($\neg$) and $m \models \neg \varphi \Leftrightarrow m \not\models \varphi$.*

We can then immediately identify a class of formulae that respect monotonicity:

**Proposition 14** *Downward persistent formulae respect monotonicity.*

**Proof:** Suppose $\varphi$ is downward persistent. Let $\alpha, \beta$ be formulae and suppose $\alpha \models_{\preceq} \beta$. Let $m$ be a minimal model of $\alpha \wedge \varphi$. Then it is also a minimal model of $\alpha$. For suppose it is not, then there exists $n \preceq m$, $n \neq m$ and $n \models \alpha$. Since $m \models \varphi$ and $\varphi$ is downward persistent, we have $n \models \varphi$. But then $n \models \alpha \wedge \varphi$ which contradicts the assumption that $m$ was a minimal model of $\alpha \wedge \varphi$. Since $m$ is a minimal model of $\alpha$ and $\alpha \models_{\preceq} \beta$, we have $m \models \beta$. We have proved that $\alpha \wedge \varphi \models_{\preceq} \beta$. Thus, $\varphi$ respects monotonicity. □

Of course, valid and unsatisfiable sentences are downward persistent. But the question is whether non-trivial downward persistent formulae exist. For the preferential logics introduced in Section 2, the answer is affirmative.

**Definition 15 (DIAM)** *Define the class of S5 formulae DIAM by:*

$$DIAM ::= M(\varphi) \mid DIAM \wedge DIAM \mid DIAM \vee DIAM \mid M(DIAM)$$

*where $\varphi$ is propositional.*

Formulae from *DIAM* essentially only contain the $M$ operator (the 'diamond' of S5, and not the 'box' operator $K$). Formulae in this class are the *only* subjective formulae (up to equivalence) which are downward persistent in Ground S5 (this was proved by Engelfriet, 1996a).

**Theorem 16** *A subjective S5 formula $\varphi$ is downward persistent in Ground S5 if and only if it is S5-equivalent to a formula in DIAM.*

So in Ground S5 there is a non-empty class of downward persistent formulae, that respect monotonicity by Proposition 14. Essentially, these formulae only say something about the ignorance of the agent. One might think that formulae from *DIAM* are completely uninteresting, and never yield any new insights in Ground S5. The converse of monotonicity for these formulae, $\alpha \wedge \varphi \models^{\text{GS5}} \beta \Rightarrow \alpha \models^{\text{GS5}} \beta$, however, does not hold, even when $\varphi$ is consistent with $\alpha$. We do not have that $Kp \vee Kq \models^{\text{GS5}} Kq$, whereas we do have that $(Kp \vee Kq) \wedge M(\neg p) \models^{\text{GS5}} Kq$ with $M(\neg p) \in DIAM$. So knowledge of ignorance can be useful.

An analogous result holds for minimal temporal epistemic logic.

**Definition 17 (TD)**

1. Define

$$TD ::= DIAM \mid TD \wedge TD \mid TD \vee TD \mid F(TD) \mid G(TD) \mid P(TD) \mid H(TD)$$

2. For two subjective TEL formulae $\varphi, \psi$:

$$\varphi \sim \psi \Leftrightarrow_{def} \text{ for all TELC models } \mathcal{M} : \mathcal{M} \models \varphi \Leftrightarrow \mathcal{M} \models \psi.$$

*TD* stands for 'temporal diamond' formulae. The following was also proved by Engelfriet (1996a).

**Theorem 18** *In MTEL, a formula $\varphi$ is downward persistent if and only if it is equivalent (in the sense of $\sim$) to a formula in TD.*

As in the case of Ground S5, these formulae express (temporal) ignorance of the agent.

**Definition 19 (Positive and universal formulae)** *A first-order predicate formula is* negative in *a predicate $P$, if all occurrences of the predicate $P$ are in the scope of an odd number of negations. A formula is* universal *if it is of the form $\forall x_1 \ldots x_n \psi$ where $\psi$ is quantifier free.*

The following result links these formulae to downward persistence in circumscription. The first is a variant of Lyndon's theorem and is folklore (we leave the details to the reader); the second result is known as the Łoś-Tarski theorem (Chang & Keisler, 1990, Theorem 3.2.2).

**Theorem 20**

1. *A first-order predicate formula $\varphi$ is downward persistent in predicate circumscription (of $P$) if and only if it is equivalent to a formula that is negative in $P$.*

2. *A first-order predicate formula $\varphi$ is downward persistent in domain circumscription if and only if it is equivalent to a universal formula.*

So downward persistent formulae in predicate circumscription essentially only say something about elements not having property $P$ (besides the other properties they mention), and downward persistent formulae in domain circumscription essentially only mention universal properties (and do not say anything about the existence of objects).

For our examples, we have shown that non-trivial classes of formulae that respect monotonicity exist. The question is whether there are more such formulae, besides those that are downward persistent. We will give a criterion that ensures that there are no more formulae that respect monotonicity.

**Definition 21 (Expressibility of preference)** *A preferential logic satisfies* expressibility of preference *if the following holds:*

$$\forall m \in Mod : \exists \varphi^m \in \mathcal{L} : \forall n \in Mod : (n \models \varphi^m \Leftrightarrow m \preceq n).$$

The formula $\varphi^m$ expresses: "I am less preferred than $m$," and describes exactly those models which are larger in the preferential ordering. The criterion of expressibility of preference poses a requirement on the expressiveness of the language, given its semantics. We will prove that in preferential logics that satisfy the condition in this definition, the downward persistent formulae are the only ones that respect monotonicity. The above condition can be generalized by taking into account equivalent models; we have not done this immediately as it makes things rather cumbersome. If whenever $n \preceq m$ and $m \equiv k$ (where $m \equiv k$ means that $m$ and $k$ satisfy the same formulae), there exists a model $l$ such that $l \equiv n$ and $l \preceq k$, then we can generalize the condition to: $\forall m \in Mod : \exists \varphi^m \in \mathcal{L} : \forall n \in Mod : (n \models \varphi^m \Leftrightarrow \exists k \in Mod : m \equiv k \ \& \ k \preceq n)$.

**Theorem 22 (Only if ...)** *For a preferential logic that satisfies expressibility of preference we have: if a formula respect monotonicity, then it is downward persistent.*

**Proof:** Suppose a formula $\varphi$ is not downward persistent, then there exist models $m$ and $n$ such that $m \models \varphi$, $n \not\models \varphi$ and $n \preceq m$. Define $\alpha = \varphi^n \wedge (\varphi \rightarrow \varphi^m)$ and $\beta = \neg\varphi$. First we claim that $\alpha \models_{\preceq} \beta$. Since $n \preceq n$, we have $n \models \varphi^n$, and as $n \not\models \varphi$ we get $n \models \alpha$. Furthermore, for any model $k$, if $k \models \alpha$ then in particular $k \models \varphi^n$ so $n \preceq k$. Therefore, $n$ is the only minimal model of $\alpha$, and since $n \not\models \varphi$, we have $n \models \beta$. On the other hand, $\alpha \wedge \varphi \not\models_{\preceq} \beta$: $n \preceq m$ so $m \models \varphi^n$ and $m \preceq m$ so $m \models \varphi^m$ from which we conclude that $m \models \alpha$ so $m \models \alpha \wedge \varphi$. Furthermore, for any model $k$, if $k \models \alpha \wedge \varphi$, then $k \models \varphi$ and $k \models \varphi \rightarrow \varphi^m$ so $k \models \varphi^m$. From this it follows that $m \preceq k$, but this means that $m$ is a (actually, the only) minimal model of $\alpha \wedge \varphi$ and $m \models \varphi$ so $m \not\models \beta$. We conclude that $\varphi$ does not respect monotonicity, since we have found formulae $\alpha$ and $\beta$ such that $\alpha \models_{\preceq} \beta$ but $\alpha \wedge \varphi \not\models_{\preceq} \beta$. $\square$

It may seem that the condition of expressibility of preference is too restrictive. However, we will see that it is useful for the examples.

**Proposition 23** *For Ground S5, MTEL and finite predicate and domain circumscription, only downward persistent formulae respect monotonicity.*

**Proof:** Remark that all of these logics satisfy Assumption 1. First consider Ground S5. Let us first take the language to be finite (that is, $P$ is finite). Take any S5 model $M$. For each propositional valuation $m$, define the formula $\alpha^m$ by $\alpha^m = \bigwedge \{p \in P \mid m \models p\} \wedge \bigwedge \{\neg p \mid p \in P, m \not\models p\}$. This is a well-defined formula since $P$ is finite. Now construct $\varphi^M = \bigwedge \{K(\neg \alpha^m) \mid m \notin M\}$, which is again a well-defined formula since $Mod(P)$ is finite. It can easily be seen that any S5 model $N$ satisfies $\varphi^M$ if and only if $M \preceq N$. So expressibility of preference is satisfied, whence Theorem 22 ensures that only downward persistent formulae respect monotonicity for this finite language. Now let $P$ be arbitrary, and suppose $\varphi$ in this language respects monotonicity. Then it is easy to see that if we restrict the language to atoms occurring in $\varphi$, it still respects monotonicity, so it is downward persistent in the restricted language. It follows easily that $\varphi$ is also downward persistent in the full language.

For MTEL, the same considerations make it sufficient to give a formula $\varphi^{\mathcal{M}}$ for a finite language only, so let us take $P$ finite. Then every S5 model $\mathcal{M}(i)$ is a finite set of propositional valuations. Since the sequence $\{\mathcal{M}(i)\}$ is decreasing with respect to set-inclusion (as $\mathcal{M}$ is conservative), there will be an index $k$ such that $\mathcal{M}(j) = \mathcal{M}(k)$ for all $j > k$. To improve readability of the formula $\varphi^{\mathcal{M}}$ we define the formulae $at_i$ to be $P^i \top \wedge H^{i+1} \bot$ (where $P^i$ stands for a sequence of $P$ operators of length $i$). It is easy to see that for any model $\mathcal{N}$ we have $(\mathcal{N}, j) \models at_i$ if and only if $j = i$. Now define:
$\varphi^{\mathcal{M}} = \bigwedge \{\Box(at_i \rightarrow \varphi^{\mathcal{M}(i)}) \mid 0 \leq i \leq k\}$, where $\varphi^{\mathcal{M}(i)}$ is the formula as defined in the case of Ground S5 for the S5 model $\mathcal{M}(i)$. It is easy to show that $\mathcal{N} \models \varphi^{\mathcal{M}}$ if and only if $\mathcal{M} \preceq \mathcal{N}$.

For finite circumscription, we need the more general definition of expressibility of preference hinted at before (in first-order logic, there may be equivalent models: different models that satisfy the same first-order formulae). Here we need not restrict the language. In predicate circumscription, the required formula $\varphi^M$ for a finite structure $M$ expresses: (i) the exact number of elements of the domain of $M$, (ii) for which of these elements $P$ holds, and (iii) for all other predicates $Q$ it expresses for which elements $Q$ holds, and for which its negation holds. In domain circumscription, the required formula $\varphi^M$ for a finite structure $M$ expresses the fact that there are (at least) as many elements as in $M$, and for each predicate $Q$, it expresses for which of these elements $Q$ holds, and for which elements its negation holds. $\square$

It is not possible to find the required formula $\varphi^M$ in (non-finite) circumscription in general: for infinite structures we are not in general able to express the number of elements, and we can not describe the entire extensions of predicates in general. Indeed, the above result does not hold for domain circumscription. It is still an open question whether it holds for predicate circumscription.

**Proposition 24** *For domain circumscription, there exists a first-order predicate formula which respects monotonicity but is not downward persistent.*

**Proof:** Consider the first-order language $\mathcal{L} = \{<, =\}$, and let $\varphi$ be a sentence stating that $<$ is a dense linear ordering without begin- or endpoint. This is a complete theory (Rabin, 1977, Theorem 4), which means that for any $\alpha \in \mathcal{L}$, either $\varphi \models \alpha$ or $\varphi \models \neg\alpha$. Now suppose $\alpha \models^{\text{DC}} \beta$. If $\varphi \models \neg\alpha$ then $\alpha \wedge \varphi$ is inconsistent, so $\alpha \wedge \varphi \models^{\text{DC}} \beta$ trivially. Otherwise we have that $\varphi \models \alpha$ so $\alpha \wedge \varphi$ is equivalent to $\varphi$. But it is easy to see that $\varphi$ does not have a minimal model, so again we have $\alpha \wedge \varphi \models^{\text{DC}} \beta$. However, $\varphi$ is not downward persistent: it holds in the real numbers, but not in the substructure of the natural numbers. □

Until now we have considered formulae that can be added to any premise, but we can also ask the question whether a formula respects monotonicity for a given, fixed premise.

**Proposition 25** *Given a preferential logic such that Mod is finite and for all $m \in Mod$ there exists $\alpha^m \in \mathcal{L}$ such that $n \models \alpha^m$ if and only if $n = m$, let $\alpha$ be a fixed formula in $\mathcal{L}$. Then we have for all $\varphi \in \mathcal{L}$:*

$$\forall \beta (\alpha \models_{\preceq} \beta \Rightarrow \alpha \wedge \varphi \models_{\preceq} \beta) \Leftrightarrow \forall m \in Mod(m \models_{\preceq} \alpha \wedge \varphi \Rightarrow m \models_{\preceq} \alpha).$$

**Proof:** The right to left direction is trivial (and does not depend on the assumption). For the other direction, suppose that $\forall \beta (\alpha \models_{\preceq} \beta \Rightarrow \alpha \wedge \varphi \models_{\preceq} \beta)$. Let $m \in Mod$ be arbitrary and suppose $m \models_{\preceq} \alpha \wedge \varphi$. Now define $\beta = \bigvee \{\alpha^n \mid n \models_{\preceq} \alpha\}$; this is a well-defined formula since $Mod$ was assumed finite. It is easy to see that $\alpha \models_{\preceq} \beta$: suppose $n \models_{\preceq} \alpha$, then $\alpha^n$ is one of the disjuncts of $\beta$, and by definition of $\alpha^n$, we have $n \models \alpha^n$, so $n \models \beta$. But the assumption now gives that $\alpha \wedge \varphi \models_{\preceq} \beta$. As $m \models_{\preceq} \alpha \wedge \varphi$, we have $m \models \beta$, so there is an $n \in Mod$ with $n \models_{\preceq} \alpha$ and $m \models \alpha^n$. But by definition of $\alpha^n$ this means that $m = n$ so $m \models_{\preceq} \alpha$. □

Proposition 25 states that a formula $\varphi$ respects monotonicity for a fixed premise $\alpha$ if and only if the minimal models of $\alpha \wedge \varphi$ are minimal models of $\alpha$. Of course the criterion on the right-hand side is hard to check; we can give another criterion, but for that, we first need the following definition (Kraus et al., 1990):

**Definition 26 (Smoothness)** *A preferential logic is called* smooth, *if the following holds:*

$$\forall \alpha \in \mathcal{L} : \forall m \in Mod : (m \models \alpha \Rightarrow \exists n \in Mod : n \preceq m \ \& \ n \models_{\preceq} \alpha).$$

This condition, which is also called *stopperedness* or *well-foundedness*, and is akin to the *limit assumption* (Lewis, 1973), forbids chains of ever-decreasing models satisfying a formula. It is one of the basic properties in the framework of Kraus et al. (1990).

**Proposition 27** *Given a smooth preferential logic, we have:* $\forall m \in Mod(m \models_{\preceq} \alpha \wedge \varphi \Rightarrow m \models_{\preceq} \alpha)$ *if and only if* $\forall m \in Mod(m \models \alpha \wedge \varphi \Rightarrow \exists n \in Mod(n \preceq m, n \models_{\preceq} \alpha \ and \ n \models \varphi))$.

The proof of this proposition is straightforward, and again it may not help much. As far as the examples are concerned, the conclusion of Proposition 25 holds for both Ground S5 and MTEL (the properties depend only on $\alpha$ and $\varphi$ so we may restrict the signature and then use the proposition). Proposition 27 holds for Ground S5 (which is smooth). From these propositions we can find some sufficient conditions. If $\varphi$ is downward persistent in the models of $\alpha$, then Proposition 25 ensures that $\varphi$ respects monotonicity with respect to $\alpha$. If $\alpha \models_\preceq \varphi$ then Proposition 27 ensures that $\varphi$ respects monotonicity with respect to $\alpha$ (but this also follows immediately with the rule of Cautious Monotonicity, which is satisfied in smooth preferential logics, Kraus et al., 1990). It seems hard to find a simple criterion necessary and sufficient for respecting monotonicity for a given premise. We leave this for further research.

## 4. Conservativity

In the previous section we have considered formulae that can always be added to a premise without invalidating any of the conclusions. In this section we will focus on the conclusions, and study formulae that, when they are concluded, can always be kept, no matter which new information is added to the premise. We will call these formulae conservative.

**Definition 28 (Conservative)** *Given a preferential logic, we say a formula $\beta$ is conservative, if*
$$\forall \alpha, \varphi : \alpha \models_\preceq \beta \Rightarrow \alpha \wedge \varphi \models_\preceq \beta.$$

We have the following result connecting upward persistent and conservative formulae, in analogy with Proposition 14.

**Proposition 29** *Given a preferential logic that is smooth, if a formula is upward persistent, it is conservative.*

**Proof:** Let $\beta$ be upward persistent in a smooth preferential logic. Now suppose $\alpha \models_\preceq \beta$. Take any model $m$ such that $m \models_\preceq \alpha \wedge \varphi$, then $m \models \alpha$ so by smoothness, there is a model $n$ with $n \preceq m$ and $n \models_\preceq \alpha$. Then, as $\alpha \models_\preceq \beta$, we have $n \models \beta$. Since $n \preceq m$ and $\beta$ is upward persistent, we have $m \models \beta$. This shows that $\alpha \wedge \varphi \models_\preceq \beta$, so $\beta$ is conservative. □

Again, we can ask if the upward persistent formulae are the only conservative formulae, and this is true under the same conditions as in the case of respecting monotonicity.

**Proposition 30 (Only if ...)** *For a preferential logic that satisfies expressibility of preference we have: if a formula is conservative, then it is upward persistent.*

**Proof:** Suppose $\beta$ is not upward persistent, then there are $n, m \in \text{Mod}$ such that $n \preceq m$, and $n \models \beta$ but $m \not\models \beta$. Now take $\alpha = \varphi^n$ and $\varphi = \varphi^m$. Then $n$ is the only minimal model of $\alpha$ and $n \models \beta$ so $\alpha \models_\preceq \beta$, but $m$ is a (actually, the only one) minimal model of $\alpha \wedge \varphi$, and $m \not\models \beta$, so $\alpha \wedge \varphi \not\models_\preceq \beta$. Thus, $\beta$ is not conservative. □

Let us first identify the upward persistent formulae for our examples. This is relatively straightforward, since we have the following elementary result.

**Proposition 31** *For any preferential logic, $\varphi$ is upward persistent if and only if $\neg\varphi$ is downward persistent.*

This gives us the following.

**Proposition 32**

1. *Define $BOX ::= K(\varphi) \mid BOX \wedge BOX \mid BOX \vee BOX \mid K(BOX)$ with $\varphi$ propositional. Then a subjective S5 formula $\varphi$ is upward persistent in Ground S5 if and only if it is S5-equivalent to a formula in BOX.*

2. *Define $TB ::= BOX \mid TB \wedge TB \mid TB \vee TB \mid F(TB) \mid G(TB) \mid P(TB) \mid H(TB)$. Then a subjective TEL formula $\varphi$ is upward persistent in MTEL if and only if it is equivalent (in the sense of $\sim$) to a formula in TB.*

3. *A first-order formula is upward persistent in predicate circumscription (of P) if and only if it is equivalent to a formula that is positive in P (meaning that all occurrences of the predicate P are in the scope of an even number of negations). A first-order formula is upward persistent in domain circumscription if and only if it is equivalent to an existential formula (a formula of the form $\exists x_1 \ldots x_n \psi$ where $\psi$ is quantifier free).*

**Proof:** Straightforward. □

In the above definition, formulae from $BOX$ essentially only contain the $K$ operator (the 'box' of S5); $TB$ stands for 'temporal box' formulae. Now let us see what Propositions 29 and 30 say about the examples. Ground S5 satisfies expressibility of preference (for a finite language) and is smooth, so the conservative formulae are exactly the upward persistent formulae, which express only knowledge (and not ignorance). This can be lifted again to an infinite language. The fact that in Ground S5, formulae that express propositional knowledge, are conservative, was already noted by Donini et al. (1997). MTEL also satisfies expressibility of preference (for a finite language), so any formula that is conservative, must be upward persistent, and must be equivalent to a formula in $TB$, expressing knowledge over time (not ignorance). This can be lifted to an infinite language. Unfortunately, MTEL is not smooth: the formula $F(Kp)$ is satisfiable, but has no minimal model. In MTEL, we have that $F(Kp) \models^{\text{MTEL}} F(Kq)$, but $F(Kp) \wedge Kp \not\models^{\text{MTEL}} F(Kq)$ ($F(Kp) \wedge Kp$ has a minimal model, in which only $p$ is known, from the first point in time onwards). This means that the formula $F(Kq)$ is not conservative, although it is upward persistent. It is easy to see that in any preferential logic, valid formulae are always conservative, but in MTEL, these are (almost) the only ones.

**Definition 33** *We call a TEL model $\mathcal{M}$ totally ignorant, if for all propositional formulae $\varphi$ we have: if $\mathcal{M} \models F(K\varphi)$ then $\varphi$ is a propositional tautology. Define the totally ignorant model $\mathcal{M}^{ti}$ by $\mathcal{M}^{ti}(i) = Mod(P)$ for all $i$.*

In a totally ignorant model, no knowledge is ever gained. *The* totally ignorant model is certainly *a* totally ignorant model, and if $P$ is finite, it is the only one.

**Proposition 34** *For MTEL, in case $P$ is infinite, we have that a formula is conservative if and only if it is true in all models. When $P$ is finite, a formula is conservative if and only if it is true in all models except possibly the totally ignorant model.*

**Proof:** We will prove that $\beta$ is conservative if and only if it is true in all models that are not totally ignorant (both when $P$ is finite and when it is infinite). First, suppose $\mathcal{M}$ is a model that is not totally ignorant, in which $\beta$ is not true. So $\mathcal{M} \models F(K\gamma)$ for some propositional formula $\gamma$ that is not a propositional tautology. If $P$ is finite, we can consider the formula $\varphi^{\mathcal{M}}$ (see the proof of Proposition 23 for the definition of $\varphi^{\mathcal{M}}$). If $P$ is infinite, one can show that $\mathcal{M}$ can be chosen in such a way that it is a 'inherently finite' model, allowing the construction of a formula $\varphi^{\mathcal{M}}$ with the same properties as for the case when $P$ is finite (using essentially the same construction). The details of this argument are left to the reader. The formula $F(K\gamma)$ does not have a minimal model ($\gamma$ must be known sometimes in the future, but this moment can always be postponed, yielding a smaller model), so $F(K\gamma) \models^{\text{MTEL}} \beta$. On the other hand, it can easily be shown that the only minimal model of $F(K\gamma) \wedge \varphi^{\mathcal{M}}$ is $\mathcal{M}$, which gives us $F(K\gamma) \wedge \varphi^{\mathcal{M}} \not\models^{\text{MTEL}} \beta$. This means that $\beta$ is not conservative.

Now suppose that $\beta$ is true in all models that are not totally ignorant, and suppose $\alpha \models^{\text{MTEL}} \beta$. Let $\mathcal{M}$ be a minimal model of $\alpha \wedge \varphi$. If $\mathcal{M}$ is not totally ignorant, then $\mathcal{M} \models \beta$. If it is totally ignorant, then also $\mathcal{M}^{ti} \models \alpha \wedge \varphi$ (it can be shown by induction that all totally ignorant models satisfy the same formulae). But then $\mathcal{M}^{ti} \models \alpha$. Since no model is preferred over $\mathcal{M}^{ti}$, this means that $\mathcal{M}^{ti} \models_{\preceq} \alpha$ so $\mathcal{M}^{ti} \models \beta$, whence $\mathcal{M} \models \beta$. It follows that $\alpha \wedge \varphi \models^{\text{MTEL}} \beta$, so $\beta$ is conservative.

Let us look at the case when $P$ is infinite. Suppose $\mathcal{M} \not\models \beta$, with $\mathcal{M}$ totally ignorant. Now take a propositional atom $p$ not occurring in $\beta$. It can easily be shown that we can find a model of $Kp$ in which $\beta$ is not satisfied. This model is of course not totally ignorant. This shows that if $\beta$ is true in all models that are not totally ignorant, then it is true in all models. In case $P$ is finite, $\mathcal{M}^{ti}$ is the only totally ignorant model. □

So in MTEL with an infinite $P$, valid formulae are the only conservative formulae. These formulae are of course upward persistent (in a trivial way), and they are equivalent to a formula in *TB*, for instance $K(\top)$. When the signature is finite, there are some extra formulae that are conservative, for example if $P = \{p, q\}$, then the formula $F(K(p \vee q) \vee K(\neg p \vee q) \vee K(p \vee \neg q) \vee K(\neg p \vee \neg q))$ is also conservative (it is true in all models except the totally ignorant one). Of course, this formula is upward persistent, and it is in *TB*.

Finite predicate and domain circumscription satisfy both expressibility of preference and smoothness, so the conservative formulae coincide with the upward persistent formulae which have the syntactic characterization of Proposition 32.

Full circumscription satisfies neither of the conditions.

**Proposition 35** *In predicate and domain circumscription there are upward persistent formulae that are not conservative.*

**Proof:** First consider predicate circumscription. Let the language consist of three predicates besides equality, namely $P$, *Succ* and $<$ (and $P$ is circumscribed). Define the formulae

Engelfriet$\alpha$ and $\varphi$ as follows:

$$\begin{aligned}\alpha = \ & \forall x \exists ! y (Succ(x,y)) \wedge & \varphi = \forall xy (Px \wedge Succ(y,x) \to Py) \\ & \forall x \exists ! y (Succ(y,x)) \wedge \\ & \forall xy (Succ(x,y) \to x < y) \wedge \\ & \forall xyz (x < y \wedge y < z \to x < z) \wedge \\ & \forall x (\neg (x < x)) \wedge \\ & \forall xy (x < y \vee y < x \vee x = y) \wedge \\ & \exists x Px \\ & \forall xy (Px \wedge Succ(x,y) \to Py) \end{aligned}$$

The intuitive meaning of $\alpha$ is that there are *Succ*-chains of elements, extending indefinitely in both directions. If $P$ occurs somewhere on such a chain, it must be true in all successors as well. A model of $\alpha$ can be made smaller (more preferred) by making $P$ false in a point and all of its predecessors (leaving it true in all successors). We will now make this argument formal. The first claim is that $\alpha$ has no $\leq_P$- minimal models. Let $M$ be a model of $\alpha$. Then there must be an $x \in dom(M)$ with $x \in P^M$. Define $A = \{x\} \cup \{y \in P^M \mid (y,x) \in <^M\}$. Let $N$ be the structure with the same domain as $M$, the same extension of *Succ* and $<$, and $P^N = P^M \setminus A$. It is straightforward to verify that $N$ is a model of $\alpha$, and that $N \leq_P M$ and $N \neq M$.

On the other hand, $\alpha \wedge \varphi$ has minimal models. Let $M$ be the structure with $dom(M) = \mathbf{Z}$ (the integers), $(a,b) \in Succ^M \Leftrightarrow b = a+1$, $(a,b) \in <^M \Leftrightarrow a < b$ in the natural ordering on the integers, and $P^M = \mathbf{Z}$. It can easily be checked that $M \models \alpha \wedge \varphi$. Now suppose $N \leq_P M$, $N \neq M$ and $N \models \alpha \wedge \varphi$. This means that $P^N \subset \mathbf{Z}$ (strict inclusion), and $P^N \neq \emptyset$ (as $N \models \exists x Px$). But then there must be $x, y \in \mathbf{Z}$ with $y = x+1$, and either $x \in P^N$ and $y \notin P^N$, or $x \notin P^N$ and $y \in P^N$, contradicting either $N \models \forall xy(Px \wedge Succ(x,y) \to Py)$ ($N \models \alpha$) or $N \models \varphi$. Therefore $M$ is a minimal model of $\alpha \wedge \varphi$.

Now define $\beta = \exists x(x \neq x)$, which is trivially upward persistent. Since $\alpha$ has no minimal models, we have $\alpha \models_P^{\text{PC}} \beta$, but $M \not\models \beta$, so $\alpha \wedge \varphi \not\models_P^{\text{PC}} \beta$. This shows that $\beta$ is not conservative.

For domain circumscription, the example is quite similar. Again take $\beta = \exists x(x \neq x)$. Now define the formulae $\alpha$ and $\varphi$ as follows:

$$\begin{aligned}\alpha = \ & \forall x \exists ! y (Succ(x,y)) \wedge & \varphi = \forall y \exists x (Succ(x,y)) \\ & \forall xy (Succ(x,y) \to x < y) \wedge \\ & \forall xyz (x < y \wedge y < z \to x < z) \wedge \\ & \forall x (\neg (x < x)) \wedge \\ & \forall xy (x < y \vee y < x \vee x = y) \wedge \\ & \forall xyz (Succ(x,z) \wedge Succ(y,z) \to x = y) \end{aligned}$$

One can now check that $\alpha$ has no $\leq_d$-minimal models, but $\alpha \wedge \varphi$ does, so the same $\beta$ is upward persistent but not conservative in domain circumscription. The details are left to the reader. □

Until now, we have looked at formulae which, once concluded, are never lost, regardless of what new information comes in, but also regardless of what the initial premise was.

However, we can also consider the situation with the premise fixed (analogously to the last part of the previous section): given a premise, which conclusions may be kept regardless of new information?

**Proposition 36** *For a preferential logic that satisfies expressibility of preference, if $\alpha \models_\preceq \beta$, then*
$$(\forall \varphi : \alpha \wedge \varphi \models_\preceq \beta) \Leftrightarrow \alpha \models \beta.$$

**Proof:** Suppose $\alpha \models_\preceq \beta$.
"$\Leftarrow$" If $\alpha \models \beta$ then for any $\varphi$ we have $\alpha \wedge \varphi \models \beta$ so $\alpha \wedge \varphi \models_\preceq \beta$.
"$\Rightarrow$" Suppose $\alpha \not\models \beta$, then there exists $m \in Mod$ such that $m \models \alpha$ but $m \not\models \beta$. Then $m \models_\preceq \alpha \wedge \varphi^m$ (!), so $\alpha \wedge \varphi^m \not\models_\preceq \beta$. □

Note that the condition $\alpha \models_\preceq \beta$ was not used in the proof; if $\alpha \not\models_\preceq \beta$ the equivalence is still true, as both sides are false. The proposition shows that the monotonic consequences of a premise are the only ones conservative with respect to this fixed premise.

**Corollary 37** *Let $\beta$ be a conservative formula for a preferential logic that satisfies expressibility of preference, then $\alpha \models_\preceq \beta \Leftrightarrow \alpha \models \beta$.*

**Proof:** If $\alpha \models \beta$ then in any preferential logic it follows that $\alpha \models_\preceq \beta$. On the other hand, if $\alpha \models_\preceq \beta$, then for any $\varphi$ we have $\alpha \wedge \varphi \models_\preceq \beta$, since $\beta$ is conservative. With Proposition 36 it follows that $\alpha \models \beta$. □

In the last two sections, we have derived a number of results on formulae that respect monotonicity and conservative formulae and the links with persistent formulae. In the next section we will discuss the impact of these results in practice.

## 5. Practical Implications

The results in this paper may improve the efficiency of theorem provers for preferential logics, depending on a number of factors. In the first place, it is important how the theorem prover is used.

Consider the situation where we have a stand-alone theorem prover which gets different (unrelated) queries. Furthermore, suppose the theorem prover is asked to prove $\alpha_1 \wedge \ldots \wedge \alpha_n \models_\preceq \beta$. Then there are at least two possibilities for using the results in the paper. First of all, suppose the preferential logic satisfies expressibility of preference. Then if $\beta$ is upward persistent, we do not have to prove $\alpha_1 \wedge \ldots \wedge \alpha_n \models_\preceq \beta$, as it is equivalent to prove $\alpha_1 \wedge \ldots \wedge \alpha_n \models \beta$ (Corollary 37). In most preferential logics, preferential entailment is harder to compute than entailment in the underlying logic.

In the second place, sometimes local reasoning is possible (which is not possible in general for non-monotonic logics): the theorem prover may derive the conclusion from part of the premise. So it may be the case that there is a $1 \leq k < n$ such that $\alpha_1 \wedge \ldots \wedge \alpha_k \models_\preceq \beta$ which is easier to verify than the original query. Then if $\alpha_{k+1}, \ldots, \alpha_n$ are downward persistent, Proposition 14 implies that $\alpha_1 \wedge \ldots \wedge \alpha_n \models_\preceq \beta$. If $\beta$ is upward persistent (and the preferential logic is smooth), Proposition 29 sanctions $\alpha_1 \wedge \ldots \wedge \alpha_n \models_\preceq \beta$. For these results to be usable in a theorem prover, we need heuristic knowledge to decide if there is a promising split of the premise into two parts $A$ and $B$. For such a split, we can then try to prove $A \models_{\preceq} \beta$ and downward persistence of $B$ or upward persistence of $\beta$. In case $\alpha_1 \wedge \ldots \wedge \alpha_k \models_{\preceq} \beta$ can not be proved, we may have to directly prove $\alpha_1 \wedge \ldots \wedge \alpha_n \models_{\preceq} \beta$ after all.

These two methods will of course only improve efficiency if the determination of persistence is easier than the original query (we will treat this question below).

The second kind of situation is when we have a theorem prover which is used by an agent which has a lot of knowledge about the world, and from time to time performs observations to increase its knowledge. Then, although sometimes the agent will need to perform revisions, we are often in the situation that (many) conclusions from a premise ($\alpha$) have been proved, and the premise is augmented by a new formula ($\varphi$). If this new formula is downward persistent, then the agent can retain all the old conclusions (and may only need to derive some new conclusions). If it is not, it can at least retain all the upward persistent conclusions (if the preferential logic is smooth). We may also try to determine if $\varphi$ is downward persistent given $\alpha$, or if $\beta$ is conservative given $\alpha$. Again, these methods only improve efficiency if it is easier to determine if $\varphi$ respects monotonicity (possibly given $\alpha$) than recomputing all old conclusions, or if it is easier to determine that $\beta$ is conservative (possibly given $\alpha$) than checking $\alpha \wedge \varphi \models_{\preceq} \beta$.

The possible efficiency improvement in both cases heavily depends on the cost of determining persistence relative to the cost of determining preferential consequence. Unfortunately, it is very hard to say anything about this issue in general. It depends on the preferential logic at hand, on the representation of the logic (syntactically, as a proof calculus, or semantically, as models with a preference relation), and on other implementation issues. For instance, it can be important how much information is retained from previous queries: whether proofs or minimal models are stored. Let us consider the examples again.

Preferential entailment in both Ground S5 and MTEL is $\prod_3^P$-complete (Engelfriet, 1996a), whereas full circumscription is undecidable (restricted versions of circumscription exist which are decidable, but still highly complex). Unfortunately, determining downward or upward persistence is not easier for these logics. We have seen that the classes of persistent formulae have syntactic representations of the form: $\varphi$ is upward/downward persistent if and only if it is equivalent to a formula in $C$, where $C$ is a (syntactic) class of formulae. Now, of course, determining *equivalence to* a formula in $C$ is as complex as determining persistence, but there may be subclasses of a class of persistent formulae, with a lower complexity. For instance, determining *membership* of $C$ is much easier, namely polynomial. The members of $C$ are persistent. So what we propose is to check membership of $C$, instead of equivalence to a member of $C$. In that case, we will miss some persistent formulae (and have to prove the original query), but this disadvantage is outweighed by the complexity advantage of checking membership. The checking of membership can be improved upon by adding some (easy) checks for equivalence to a formula in $C$. For instance, in Ground S5, if we consider, for a formula $\varphi$, for each propositional sub-formula, the nearest $K$ operator in which scope it lies, then if all of these $K$ operators are in the scope of an odd number of negations, we can conclude that $\varphi$ is downward persistent. The formula $\neg K(q \vee Kp)$, for example, satisfies this condition, and although it is not a member of $DIAM$, it is equivalent to $\neg Kq \wedge \neg Kp \in DIAM$. This check is obviously polynomial.

Given a preferential logic, the designer of a theorem prover could proceed as follows. First, syntactic classes of formulae that are downward and upward persistent have to be identified. For Ground S5, MTEL and predicate and domain circumscription, these can be found in Definitions 15, 17, 19 and Proposition 32. For other preferential logics, if such classes are trivial (they may, for instance, only include tautologies and contradictions), then the usefulness of the results is limited. Otherwise, the theorem prover could work as follows. Given a query of the form $\alpha \wedge \varphi \models_\preceq \beta$, first it is checked if $\varphi$ belongs to the syntactic class of downward persistent formulae or if $\beta$ belongs to the syntactic class of upward persistent formulae (this latter test should only be performed if the preferential logic is smooth). If $\beta$ is conservative and the logic satisfies expressibility of preference, it tries to prove $\alpha \wedge \varphi \models \beta$ (this usually has a lower complexity than the original query; for Ground S5 and MTEL, monotonic consequence is NP-complete, Engelfriet, 1996a). The answer of this query is the answer to the original query (see Corollary 37). Otherwise, if $\varphi$ belongs to the syntactic class of downward persistent formulae or if $\beta$ belongs to the syntactic class of upward persistent formulae (but the logic does not satisfy expressibility of preference), then the theorem prover tries to prove $\alpha \models_\preceq \beta$. If this succeeds, it outputs yes. Otherwise, it will try to answer the original query directly.

As stated before, the practical savings in part depend on representation and implementation aspects. It also depends on the application domain and use of the theorem prover: if formulae in these syntactic classes occur often, the efficiency improvement is higher than if they are infrequent.

## 6. Conclusions and Further Research

We have looked at restrictions of monotonicity in preferential logics. While monotonicity does not hold in general, we can identify (in general non-trivial) classes of formulae for which restricted versions of monotonicity holds. This may make (nonmonotonic) deduction in preferential logics easier, since we may sometimes keep conclusions, or identify which conclusions may be kept, when adding new information to a premise. The results in this paper may lead to more efficient implementations of preferential logics. Experimenting with theorem provers which use these results is necessary in order to determine the efficiency improvement in practice.

It would be nice to find a better characterization of formulae that can be added to a given, fixed premise without destroying conclusions.

Syntactic characterizations of persistent formulae were given for a number of example preferential logics, but we would like to have a result for broader classes of preferential logics, such as the class of ground nonmonotonic modal logics (Donini et al., 1997).

## Acknowledgements

I would like to thank Riccardo Rosati for stimulating discussions and worthwhile suggestions. Also, I would like to thank Pascal van Eck, Jan Treur, Rineke Verbrugge, Elles de Vries and Wieke de Vries for reading and commenting on earlier versions of this doc-ument. Furthermore, I would like to thank Heinrich Herre for suggesting the example in Proposition 24. The anonymous referees helped to improve the paper.